\documentclass[conference,compsoc]{IEEEtran}
\ifCLASSOPTIONcompsoc
  \usepackage[nocompress]{cite}
\else
  \usepackage{cite}
\fi
\ifCLASSINFOpdf
\else
\fi
\usepackage{amsmath,amssymb,amsfonts}
\usepackage{longtable}
\usepackage{graphicx}
\usepackage{textcomp}
\usepackage{xcolor}
\usepackage{booktabs}
\usepackage{multirow}
\usepackage{color}
\usepackage{url}
\usepackage{cite}
\begin{document}
%
\title{A Multiscale Image Denoising Algorithm Based On Dilated Residual Convolution Network}

\author{\IEEEauthorblockN{Chang Liu}
\IEEEauthorblockA{College of Computer Science\\
Chongqing University\\
Chongqing, China\\
Email: changliu37@cqu.edu.cn}
\and
\IEEEauthorblockN{Zhaowei Shang}
\IEEEauthorblockA{College of Computer Science\\
Chongqing University\\
Chongqing, China\\
Email: szw@cqu.edu.cn}
\and
\IEEEauthorblockN{Anyong Qin}
\IEEEauthorblockA{College of Computer Science\\
Chongqing University\\
Chongqing, China\\
Email: ayqin@cqu.edu.cn}}


%


\maketitle

\begin{abstract}
Image denoising is a classical problem in low level computer vision. Model-based optimization methods and deep learning approaches have been the two main strategies for solving the problem. Model-based optimization methods are flexible for handling different inverse problems but are usually time-consuming. In contrast, deep learning methods have fast testing speed but the performance of these CNNs is still inferior. To address this issue, here we propose a novel deep residual learning model that combines the dilated residual convolution and multi-scale convolution groups. Due to the complex patterns and structures of inside an image, the multiscale convolution group is utilized to learn those patterns and enlarge the receptive field. Specifically, the residual connection and batch normalization are utilized to speed up the training process and maintain the denoising performance. In order to decrease the gridding artifacts, we integrate the hybrid dilated convolution design into our model. To this end, this paper aims to train a lightweight and effective denoiser based on multiscale convolution group. Experimental results have demonstrated that the enhanced denoiser can not only achieve promising denoising results, but also become a strong competitor in practical application.
\end{abstract}

%
\IEEEpeerreviewmaketitle
\section{Introduction}
Image denoising is a classical yet still active theme in computer vision. It has become an essential and indispensable step in many image processing applications. In recent years, various algorithms have been proposed, which include nonlocal self-similarity(NSS) models~\cite{IEEEexample:Mairal2010Non}, sparse representation models~\cite{IEEEexample:Dong2013Nonlocally}, and deep learning approaches~\cite{IEEEexample:Zhang2017Beyond}~\cite{IEEEexample:Kim2017Deeply}~\cite{IEEEexample:Chen2017Trainable}. Among them, BM3D~\cite{IEEEexample:Dabov2007Image}, WNNM~\cite{IEEEexample:Gu2014Weighted} and CSF~\cite{IEEEexample:Schmidt2014Shrinkage} are considered as the stats-of-the-art methods in non-depth learning approaches. NSS models like BM3D provide high image quality and efficiency, and they are very effective in Gaussian denoising with known noise level. In recent, many state-of-the-art CNN algorithms like IRCNN~\cite{IEEEexample:Zhang2017Learning} outperform the non-local and collaboration filtering approaches. The deep CNNs are trained to learn the image prior, which shows that CNN algorithms have a strong ability to fit the structure and pattern inside the image.

Since the deep learning methods have achieved massive success in classification~\cite{IEEEexample:Krizhevsky2012ImageNet} as well as other computer vision problems~\cite{IEEEexample:Ronneberger2015U}~\cite{IEEEexample:Wang2017Understanding}. A lot of CNN algorithms have been proposed in
image denoising. Aiming at image restoration task, it is important to use the prior information properly. Many prior-based approaches like WNNM involve a complex optimization problem in the inference stage, this leads to achieve high performance hardly without sacrificing computation efficiency. To overcome the limitation of prior-based methods, several discriminative learning methods have been developed to learn image prior models in the inference procedure. This kind of models can get rid of the iterative optimization procedure in the test phase. Inspired by the work of~\cite{IEEEexample:Zhang2017Beyond}~\cite{IEEEexample:Zhang2017Learning}, this task of image denoising can be seen as a Maximum A Posteriori(MAP) problem from the Bayesian perspective. The deep CNN can be used to learn the prior as a denoiser. The motivation of this work is whether we can increase the prior from the view of convolution itself. The work of ~\cite{IEEEexample:Szegedy2015Going} shows the design of multi-scale convolutions which can help to extract more features from the previous layer. Reference ~\cite{IEEEexample:Yu2017Dilated} introduced the work of dilated convolution, which can enlarge the receptive field and keep the amount of calculation. Subsequently,  the work of~\cite{IEEEexample:Wang2017Understanding} shows that dilated residual network can perform better than the residual network in the image classification. In the image denoising task, due to the little difference between adjacent pixels, the dilated convolution can bring more discrepancy information from the front layer to the back layer. In addition, it could increase the generalization ability of the model and require no extra computation cost. The design of residual module can accelerate the whole training process and prevent the gradient vanishing.

This paper proposes an enhanced denoiser based on the residual dilated convolutional neural network. Inspired by the residual learning insight proposed by ~\cite{IEEEexample:Zhang2017Beyond}, we modified the dilated residual network based on residual learning. We treat image denoising as a plain discriminative learning problem. So CNN is usually utilized to separate the noisy. The reasons of using CNN can be summarized as the following. First, CNN with very deep architecture is effective in exploiting image characteristics and increasing the capacity and flexibility of the model. Second, a lot of progress have been made in CNN training methods, including parametric rectifier linear unit(PReLU)~\cite{IEEEexample:He2015Delving}, batch normalization and residual architecture. These methods can speed up the training process and improve the denoising performance. Third, there are many tools and libraries to support parallel computation for CNN on GPU, which can give a significant improvement on runtime performance. Contrary to existing various residual networks, we use multi-scale convolution to extract more information from the original image and hybrid dilated convolution module to avoid the gridding effect. In the experiment, we compare several stat-of-the-art methods, such as BM3D, IRCNN, DnCNN~\cite{IEEEexample:Zhang2017Beyond} and FFDnet~\cite{IEEEexample:Zhang2017FFDNet}. For gaussian denoising, our result shows that the proposed enhanced denoiser can make the processed image better with only half parameters of other CNN methods. And the proposed model has a competitive run time performance.

In summary, this paper has the following two main contributions:

Firstly, we proposed a lightweight and effective image denoiser based on multi-scale convolution group. The experiments shows that our proposed model can achieve better performance and speed over the current stat-of-the-art methods.

Secondly, we shows the proposed network can handle both gray and color image denosing robustly without the increment of parameters.

The rest of this paper is organized as follows. Section 2 gives a review of recent image denoising approaches. Section 3 formally describe our research problem and method in detail. Section 4 presents the experimental results of proposed method and the comparison with one baseline model and five different model. Finally, we conclude in section 5.
\begin{figure*}[t]
\centering
\includegraphics[width=15cm]{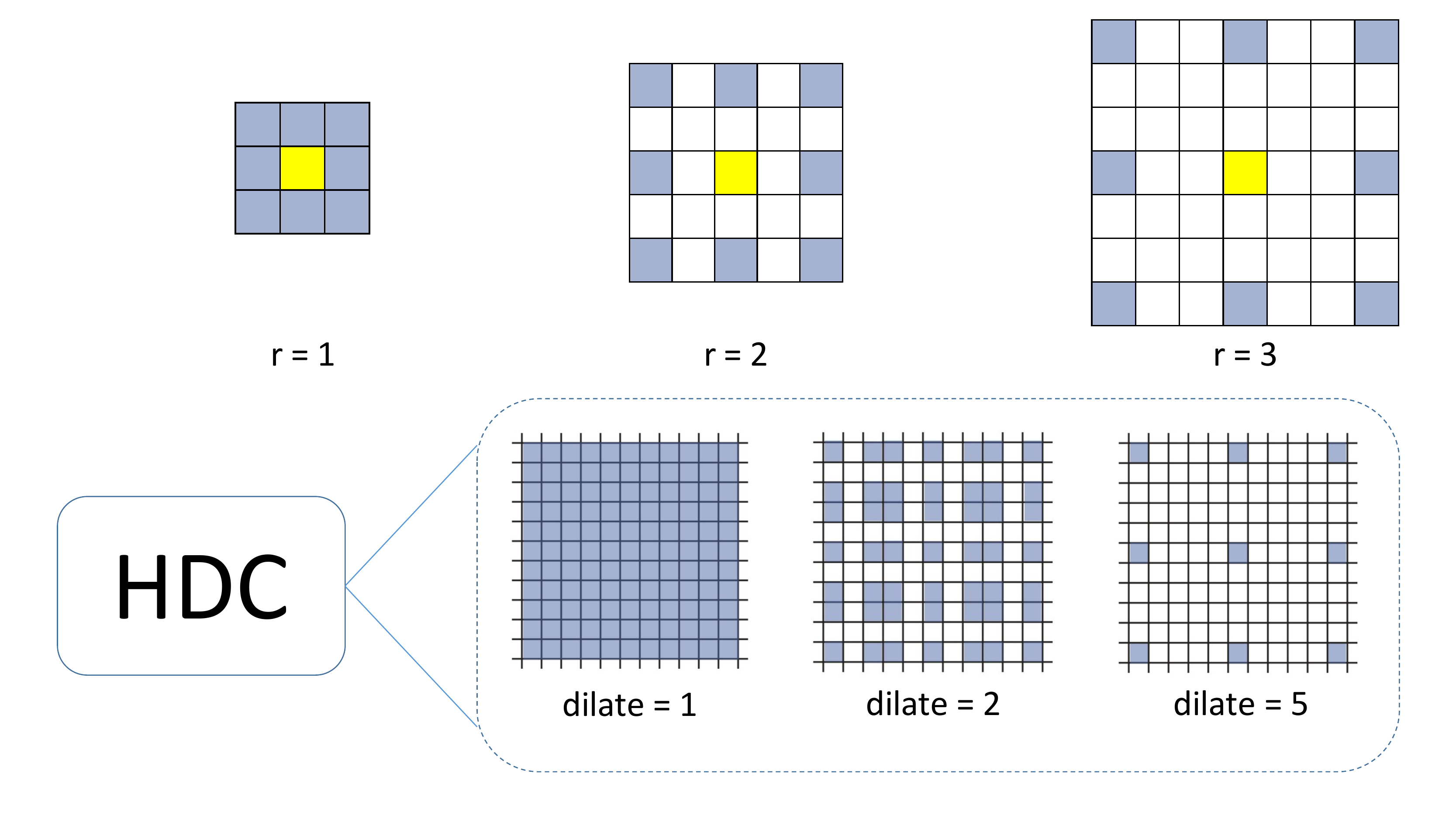}
\caption{Illustration of the dilate convolution and Hybrid Dilated Convolution(HDC) architecture. The pixels(marked in lavender) contributes to the calculation of the center pixels(marked in yellow) through three different dilation rate $r$ with $3\times3$ convolution filters. HDC is consisted of three convolution layers which have dilation rate of $r = 1,2,5,$ respectively}
\label{fig1}
\end{figure*}

\section{Related Work}
Here, we provide a brief review of image denoising methods. Harmeling et al.~\cite{IEEEexample:Harmeling2012Image} was firstly to apply multi-layer perception(MLP) for image denoising task, which image patches and large image databases were utilized to achieve excellent results. In ~\cite{IEEEexample:Chen2017Trainable}, a trainable nonlinear reaction diffusion(TNRD) model was proposed and all the parameters can be simultaneously learned from training data through a loss based approach. It can be expressed as a feed-forward deep network by unfolding a fixed number of gradient inference steps. DeepAM~\cite{IEEEexample:Kim2017Deeply} is consisted of two steps, proximal mapping and end continuation. It is the regularization-based approach for image restoration, which enables the CNN to operate as a prior or regularizer in the alternating minimization(AM) algorithm. IRCNN~\cite{IEEEexample:Zhang2017Learning} uses the HQS framework to show that CNN denoiser can bring strong image prior into model-based optimization methods. All the above methods have shown that the decouple of the fidelity term and regularization term can enable a wide variety of existing denoising models to solve image denoising task.

Residual learning has multiple realization. The first approach is using a skipped connection from a certain layer to another layer during forward and backward propagations. This was firstly introduced by ~\cite{IEEEexample:He2015Deep} to solve the gradient vanishing when training very deep architecture in image classification. In low-level computer vision problems, implemented a residual module within three convolution block by a skipped connection. Another residual implementation is transforming the label data into the difference between the input data and the clean data. The residual learning~\cite{IEEEexample:Zhang2017Beyond} can not only speed up the training, but also make the weights of network sparser.

Dilated convolution was originally developed for wavelet decomposition~\cite{IEEEexample:Holschneider1989A}. The main idea of dilated convolution is to increase the image resolution by inserting "holes" between pixels. The dilated convolution enables dense feature extraction in deep CNNs and enlarges the field of convolutional kernel. Chen et al. ~\cite{IEEEexample:Chen2018DeepLab} designs an atrous spatial pyramid pooling(ASPP) scheme to capture multi-scale objects and context information by using multiple dilated convolution. In image denoising, Wang et al.~\cite{IEEEexample:Wang2017Dilated} proposed an approach to calculate receptive field size when dilated convolution is included.

\section{Method}
\begin{figure*}[htbp]
\centering
\includegraphics[width=16cm]{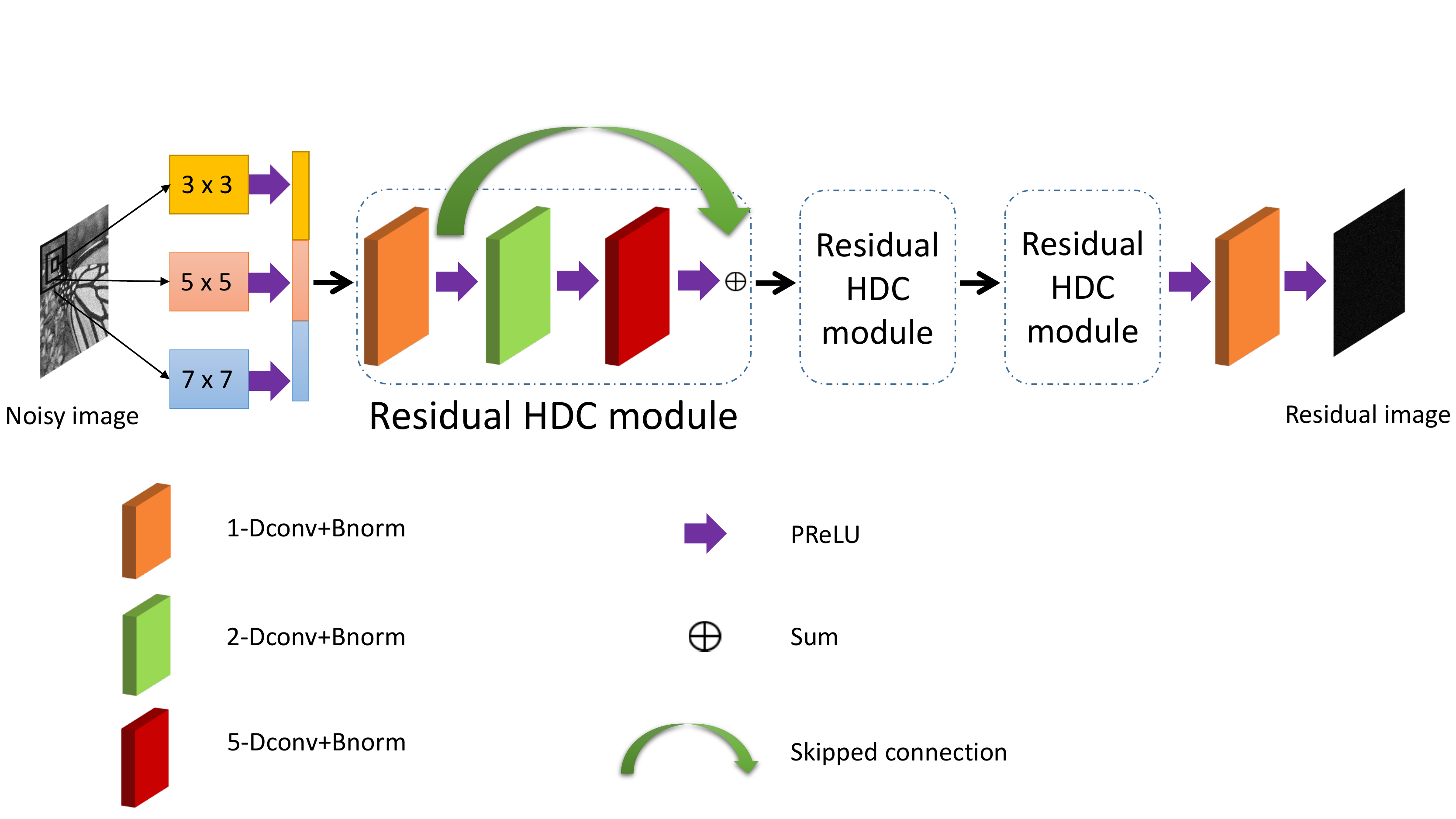}
\caption{The architecture of the proposed denoiser network. Note that the residual HDC module consists of the shortcuts and HDC design block. "s-Dconv" represents dilated convolution, here $s = 1,2$ and $5$; "Bnorm" represents batch normalization~\cite{IEEEexample:Ioffe2015Batch}; "PRelu" is the activation function~\cite{IEEEexample:He2015Delving}. }
\label{fig2}
\end{figure*}
\subsection{Dilated filter}
Dilated filter was introduced to enlarge receptive field. The context information can facilitate the reconstruction of the corrupted pixels in image denoising. In CNN, there are two basic ways to enlarge the receptive field to capture the context information. We can either increase the filter size or the depth. According to the existing network design~\cite{IEEEexample:Szegedy2015Going}~\cite{IEEEexample:He2015Deep}, using the $3\times3$ filter with a large depth is a popular way. In this paper, we use the dilated filter and keep the merits of traditional $3\times3$ convolution. In Fig.~\ref{fig1}, the image filtered by different dilate rate shows the different receptive field. A dilated filter with dilation factor $s$ can be simply interpreted as a sparse filter of size $(2s+1)\times(2s+1)$. For kernel size $K = 3$ and dilate rate $r = 2$, only 9 entries of fixed positions can be non-zeros. But the use of dilated convolutions may cause gridding artifacts~\cite{IEEEexample:Yu2017Dilated}. It occurs when a feature map has higher-frequency content than the sampling rate of the dilated convolution. And we notice that the hybrid dilated convolution(HDC) proposed by ~\cite{IEEEexample:Wang2017Understanding} addressed this issue theoretically. Suppose $N$ convolutional layers with kernel size $K \times K$ have dilation rates of $[r_{1},r_{2},\ldots,r_{n}]$, the HDC is going to let the final size of receptive field cover a square region without any holes or missing edges. So the maximum distance between two nonzero values can be defined as
\begin{equation}\label{7}
 M_{i} = max[M_{i+1} - 2r_{i},M_{i+1} - 2(M_{i+1} - r_{i}), r_{i}]
\end{equation}
where the design goal is to let $M_{2}\leq K$ with $M_{n} = r_{n}$. For example, for kernel size $K = 3$, $r = [1,2,5]$ pattern works as $M_{2} = 2$; however, an $r = [1,2,9]$ pattern does not work as $M_{2} = 5$. The benefit of HDC is that it can naturally integrated with the original layers of network, without adding extra modules.

The HDC can make the better use of receptive field information. 1 dilate convolution and 5 dilate convolution will extract features information at different level. In other word, the dilated convolution can extract information at different scale.

\subsection{Multiscale convolution group}
Multiscale extraction for image feature is a common technique in solving computer vision problems. Multiscale extraction can make use of feature maps in different levels. In deep CNNs, the $3\times3$ kernel and $5\times5$ kernel can extract different scales of features. The addition of the multiscale structure increases the width of network , on the other hand, improves the generalization of network. Inspired by the Inception module~\cite{IEEEexample:Szegedy2014Going}, we proposed the multiscale convolution group. Inception module consists of the pooling layer and $1\times1$ convolution. For image denoising task, due to the unchanged size of output image, we usually remove the pooling layer. We apply three scale filters with different numbers. The number of each of the three filter is 12, 20 , 32. The reason is that the sum of the feature map is 64 and this combination achieves the balance between feature extraction and parameters. Differing from the Inception module, we concatenate the feature maps directly. It can significantly reduce the parameters. Considering of the computation cost, we only use the multiscale module in the first layer.

\subsection{Architecture}
Our proposed network structure illustrated in fig.~\ref{fig2} is inspired by ~\cite{IEEEexample:Zhang2017Learning} and ~\cite{IEEEexample:Wang2017Understanding}. It consists of eleven layers with different dilated rate convolution. The first layer is the multiscale convolution group. The residual connection starts from the second layer. The skipped connection is used for training deep network because it is beneficial for alleviating the gradient vanishing problem~\cite{IEEEexample:Mao2016Image}~\cite{IEEEexample:He2016Identity}. Another advantage is the residual module can make the weights of network sparse, which can reduce the inference computation time. Each dilated convolution block will be followed by  batch normalization~\cite{IEEEexample:Ioffe2015Batch} layers and parametric rectifield linear unit(PRelu)~\cite{IEEEexample:He2015Delving}. Such network design techniques have been widely used in recent CNN architecture. In particular, it has been pointed out that this kind of combination can not only enables fast and stable training but also tends to better result in Gaussian denoising task~\cite{IEEEexample:Zhang2017Learning}. The PReLU and BN layer can accurate the convergence of network. Image denoising is going to recover a clean image from a noisy observation. We consider the noisy image $y$ can be expressed as $y = x + G$, where $x$ stands for the clean image, and $G$ is the unknown Gaussian noise distribution. Owing to residual learning strategy applied, the labels are obtained by calculating the difference between the input image and the clean image. So the output of our network is the residual image, which is the prediction of the noise distribution.

\section{Experiments}

It is widely acknowledged that convolution neural networks generally benefits from the giant training datasets. However, due to the limitation of computer resources, we only used 400 images of size $180\times180$ for gray-scale denoising experiments. According to ~\cite{IEEEexample:Zhang2017Learning}, using more dataset dose not improve the PSNR results of BSD68 dataset. As for the color image denoising experiment, 400 selected images from validation set of ImageNet database~\cite{IEEEexample:Deng2009ImageNet} are the training datasets and we crop the images into small patches of size $45 \times 45$ and select $N = 128 \times 2000$ patches for training.

\subsection{Training details}
Due to the characteristic of Gaussian convolution, the output image may produce boundary artifacts. So we apply zero padding strategy and use small patches to tackle with this problem. The number of feature maps in each layer is 64. And the depth of network is set to eleven which is kind of lightweight framework. The patch size of input images is $45\times45$, and we use date augmented techniques to increase the diversity of the training data. The Adam solver ~\cite{IEEEexample:Kingma2014Adam} is applied to optimize the network parameters $\Theta$. The learning rate is started from $1e-3$ and then fixed to $1e-4$. The learning rate is decreasing 10 times after 60 epoches. The hyperparameters of Adam is set the default setting. The mini-batch size we use is 64, which can balance the memory usage and effects well. The network parameters are initialized using the Xavier initialization~\cite{IEEEexample:He2015Delving}. And our denoiser models are trained with the MatConvNet package~\cite{IEEEexample:Vedaldi2014MatConvNet}. The environment we use is under Matlab(2016b) software and an Nvidia 1080Ti GPU. And we trained 100 epoches to get the result, it almost takes half a day to train a denoised model with the specific noise level. 400 images of publicly available Berkeley segmentation(BSD500)~\cite{IEEEexample:Chen2017Trainable} and Urban 100 datasets~\cite{IEEEexample:Huang2015Single} are used for training the Gaussian denoising model. In addition, we generated 3200 images by using data augmentation via image rotation, cropping and flipping. For the test data set, Set12 and BSD68 are used. Note that all those image are widely used for the evaluation of denoising models and they are not included in the training datesets. For training and validation, Gaussian noises with $\sigma = 15, 25, 50$ are added to verify the effect of our denoised model.

\subsection{denoising results}

We compared the proposed denoiser with several state-of-the-art denoising methods, including two model-based optimization methods BM3D and WNNM, one discriminative learning method TNRD, and three deep learning methods included IRCNN, DnCNN, FFDnet. Fig. ~\ref{fig3} shows the visual results with details of different methods. It can be seen that both BM3D and WNNM tend to produce over-smooth textures. TRND can preserve fine details and sharp edges, but it seems that artifacts in the smooth region are generated. The three deep learning methods and the proposed method can have a pleasure result in the smooth region. It is clearly that the proposed method can preserve better texture than the other methods, such as the region above the balcony fence. Another comparison results of different methods in fig.~\ref{fig4} show that our method reaches a better visual result.

\begin{table*}[t]
\caption{Performance comparison in terms of average PSNR(dB) results for BSD68 dataset, the best results are highlighted in bold. }
\centering
\begin{tabular}{|c|c|c|c|c|c|c|c|c|}
\hline
Methods&BM3D&WNNM&TNRD&IRCNN&DDRN&DnCNN&FFDNet&Proposed\\
\hline
$\sigma=15$&31.075&31.371&31.422&31.629&31.682&31.718&31.631&\textbf{31.751}\\
\hline
$\sigma=25$&28.568&28.834&28.923&29.145&29.181&29.228&29.189&\textbf{29.258}\\
\hline
$\sigma=50$&25.616&25.874&25.971&26.185&29.213&26.231&26.289&\textbf{26.323}\\
\hline
$\sigma=75$&24.212&24.401&-&24.591&24.617&24.641&24.788&\textbf{24.793}\\
\hline
\end{tabular}

\label{table1}
\end{table*}

In order to show the capacity of the proposed model, we do the quantitative and qualitative evaluation on 2 widely used testing data sets. The average PSNR results of different results on the BSD68 and Set12 are shown in table.\ref{table1} and table.\ref{table2}. BSD68 consists of 68 gray images, which has diverse images. We can have the following observation. Firstly, the proposed method can achieve the best average PSNR result than those competing methods on BSD68 data sets. Compared to the benchmark method BM3D on BSD68, the WNNM and TNRD have a notable gain of between 0.3dB and 0.35 dB. The method IRCNN can have a PSNR gain of nearly 0.55dB. In contrast, our proposed model can outperform BM3D nearly 0.7dB on all the three noise levels. Secondly, the proposed method is better than DnCNN and FFDnet when the noise level is below 75. This result shows that the proposed method has the better trade-off between receptive field size and modeling capacity.

Table.\ref{table2} lists the PSNR results of different methods on the 12 test images. The best two PSNR result for each image with each level is highlighted in red and blue color. It can be seen that the proposed method can achieve the top two PSNR values on most of the test images. For the average PSNR values, the proposed method has best performance among all the methods in $\sigma=15, 25$ noisy. And it is less efficient than FFDnet in $\sigma=50$ noisy. This is because FFDnet can outperforms the other methods on image "House" and "Barbara", which this two images have rich amount of repetitive structures.

\begin{figure*}[htbp]
\centering
\includegraphics[width=16cm]{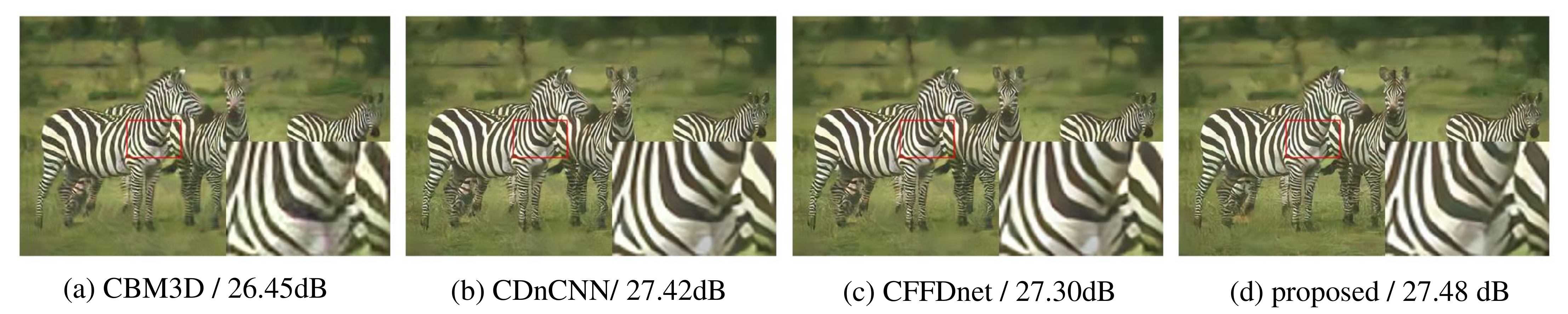}
\caption{Color image denoising results of one image from the CBSD68 dataset with noise level 50}
\label{fig7}
\end{figure*}

For color image denoising, we use the same network parameters. The only difference is the input tensor becomes $45\times45\times3$. The visual comparisons are shown in fig.\ref{fig6} and fig.\ref{fig7}. It is obviously that CBM3D generates false color artifacts in some region while the proposed model can recover the image with more natural color and texture structure, like more sharp edges. In addition, table.\ref{table4} shows that proposed model can outperform the benchmark method CBM3D among three noise level. In the meantime, the proposed method is more effective than three deep CNN methods in the color BSD68 dataset.

\begin{table}[htbp]
\caption{Performance comparison in terms of average PSNR(dB) results for color BSD68 dataset }
\centering
\begin{tabular}{|c|c|c|c|}
\hline
Methods&$\sigma = 15$&$\sigma = 25$&$\sigma = 50$\\
\hline
CBM3D&33.52&30.71&27.38\\
\hline
CDDRN&33.93&31.24&27.93\\
\hline
CDnCNN&33.89&31.23&27.92\\
\hline
CFFDnet&33.87&31.21&27.96\\
\hline
Proposed&\textbf{34.10}&\textbf{31.43}&\textbf{28.09}\\
\hline
\end{tabular}
\label{table4}
\end{table}

We give a brief calculation about the amount of parameters. Note that the values are different for gray and color image denoising due to the different network depth. For instance, DnCNN uses 17 convolution layers for gray image denosing and 20 for color image denoising, whereas FFDnet takes 15 for gray and 12 for color. In addition, FFDnet set 64 channels for gray image and 96 channels for color image. However, the proposed method can outperform the other method without the increment of the depth in color image denoising. It indicates that our model is more robust without sacrificing the computing resource.

\begin{table}[htbp]
\caption{The amount of parameters for three different method. The color denoiser contains more parameters due to the deeper architecture}
\centering
\begin{tabular}{|c|c|c|c|}
\hline
Methods&gray/param&color/param\\
\hline
DnCNN&5.6 $\times$ 10$^{5}$&6.7 $\times$ 10$^{5}$\\
\hline
FFDnet&5.5 $\times$ 10$^{5}$&8.3 $\times$ 10$^{5}$\\
\hline
Proposed&3.3 $\times$ 10$^{5}$&3.4 $\times$ 10$^{5}$\\
\hline
\end{tabular}
\label{table5}
\end{table}

We also compare the computation time to check the applicability of the proposed method. BM3D and TNRD are utilized to be the comparison due to their potential value in practical applications. We use the Nvidia cuDNN-v6 deep learning library to accelerate the GPU computation and we do not consider the memory transfer time between CPU and GPU. Since both the proposed denoiser and TNRD support parallel computation on GPU, we also provide the GPU runtime. Table.~\ref{table3} lists run time comparison of different methods for denoising images of size $256\times256$, $512\times512$ and $1024\times1024$. For each kind of tests, we run several times to get the average runtime. We can see that the proposed method is very competitive in both CPU and GPU computation. Such a good performance over the BM3D is properly attributed to the following reasons. First, the $3\times3$ convolution and PRelu activation function are simple effective and efficient. Second, batch normalization is adopted, which is beneficial to Gaussian denoising. Third, residual architecture can not only accelerate the inference time of deep network, but also have a larger model capacity.

\begin{table}[htbp]
\caption{Run time of different methods on gray images of different size with noise level 25}
\centering
\begin{tabular}{|c|c|c|c|c|}
\hline
Size&Device& 256$\times$256 & 512$\times$512 & 1024$\times$1024 \\
\hline
\multirow{2}{*}{BM3D}&CPU &0.69& 0.52& 0.371\\
&GPU&-& - & - \\
\hline
\multirow{2}{*}{DnCNN}&CPU&2.14&8.62&32.10\\
&GPU&0.018& 0.046 & 0.135\\
\hline
\multirow{2}{*}{FFDnet}&CPU&0.44&1.81&7.32\\
&GPU&0.008& 0.016 & 0.046\\
\hline
\multirow{2}{*}{Proposed}&CPU&0.41&1.62&4.68\\
&GPU&0.004& 0.009 & 0.038\\
\hline
\end{tabular}
\label{table3}
\end{table}

\begin{figure*}[htbp]
\centering
\includegraphics[width=18cm]{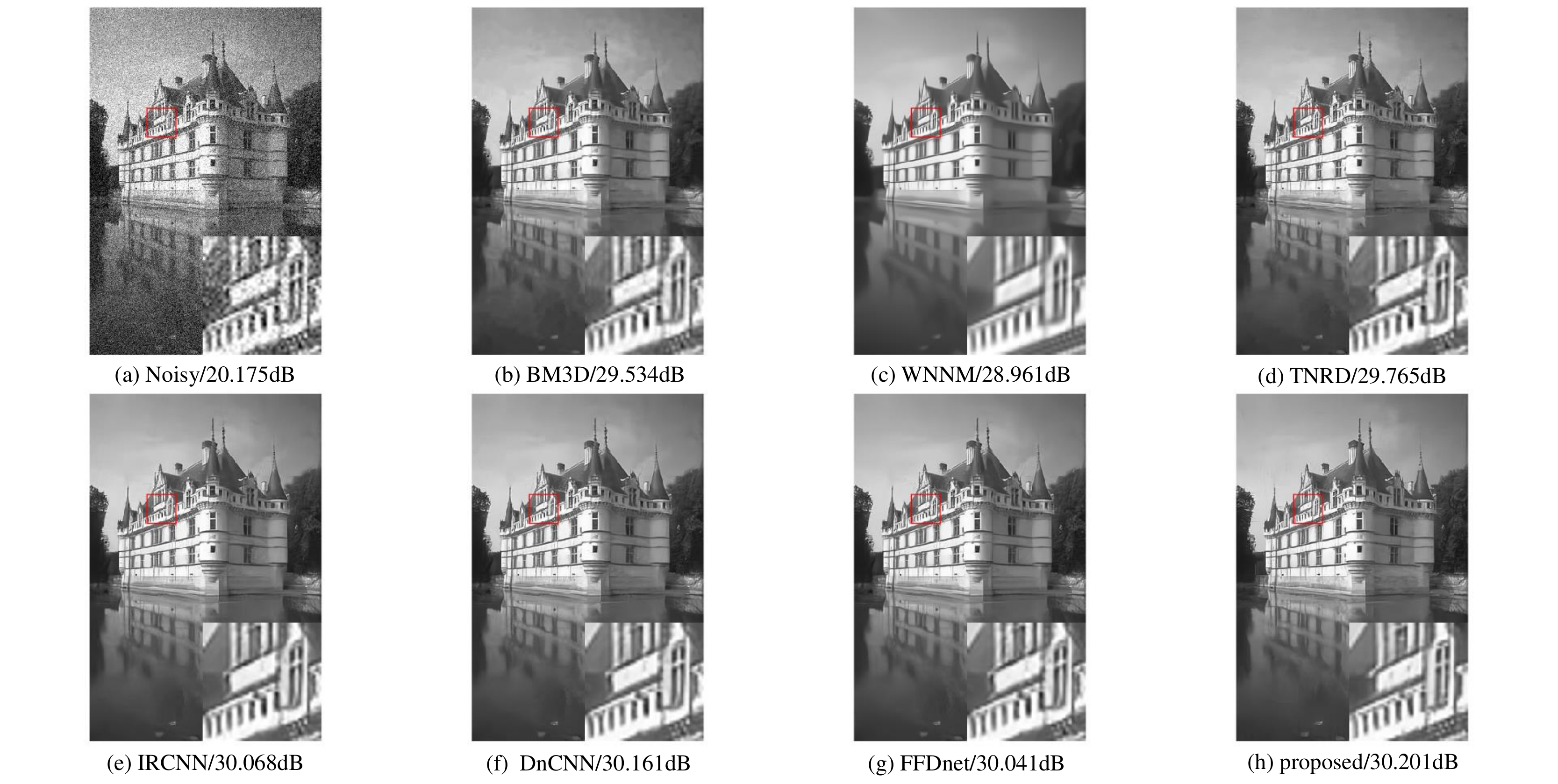}
\caption{Denoising results of one image from BSD68 with noise level 25}
\label{fig3}
\end{figure*}

\begin{table*}[htbp]
\caption{The PSNR(dB) performance of different methods on 12 widely used testing images. }
\centering
\resizebox{\textwidth}{45mm}
{
\begin{tabular}{|c|c|c|c|c|c|c|c|c|c|c|c|c|c|}
\hline
Images&C.man&House&Peppers&Starfish&Monar.&Airpl.&Parrot&Lena&Barbara&Boat&Man&Couple&Average\\
\hline
Noise Level&\multicolumn{13}{|c|}{$\sigma = 15$}\\
\hline
BM3D&31.915&34.944&32.701&31.146&31.859&31.076&31.376&34.271&\textcolor{blue}{33.114}&32.140&31.929&32.108&32.372\\
\hline
WNNM&32.168&\textcolor{red}{35.129}&32.987&31.825&32.712&31.387&31.621&34.273&\textcolor{red}{33.598}&32.269&32.115&32.172&32.688\\
\hline
TNRD&32.154&34.541&33.069&31.702&32.611&31.484&31.709&34.289&32.098&32.143&32.178&32.065&32.502\\
\hline
IRCNN&32.539&34.886&\textcolor{red}{33.314}&32.021&32.816&\textcolor{blue}{31.698}&\textcolor{blue}{31.839}&34.528&32.431&32.342&32.398&32.401&32.769\\
\hline
DnCNN&\textcolor{blue}{32.611}&34.972&\textcolor{blue}{33.297}&\textcolor{red}{32.197}&\textcolor{blue}{33.087}&31.696&31.831&\textcolor{blue}{34.621}&32.638&\textcolor{blue}{32.416}&\textcolor{blue}{32.451}&\textcolor{red}{32.471}&\textcolor{blue}{32.857}\\
\hline
FFDNet&32.417&35.005&33.102&32.021&32.768&31.587&31.768&\textcolor{red}{34.632}&32.501&32.348&32.402&32.447&32.749\\
\hline
Proposed&\textcolor{red}{32.633}&\textcolor{blue}{35.006}&33.261&\textcolor{blue}{32.161}&\textcolor{red}{33.238}&\textcolor{red}{31.722}&\textcolor{red}{31.908}&34.598&32.586&\textcolor{red}{32.443}&\textcolor{red}{32.452}&\textcolor{blue}{32.454}&\textcolor{red}{32.872}\\
\hline
Noise Level&\multicolumn{13}{|c|}{$\sigma = 25$}\\
\hline
BM3D&29.452&32.864&30.161&28.561&29.254&28.427&28.934&32.077&\textcolor{blue}{30.717}&29.908&29.615&29.719&29.969\\
\hline
WNNM&29.642&\textcolor{blue}{33.221}&30.421&29.031&29.842&28.693&29.152&32.243&\textcolor{red}{31.239}&30.031&29.752&29.821&30.257\\
\hline
TNRD&29.634&32.504&30.392&28.952&29.978&28.928&29.245&32.016&29.327&29.867&29.824&29.703&30.031\\
\hline
IRCNN&30.083&33.056&30.825&29.267&30.085&29.087&\textcolor{blue}{29.461}&32.421&29.924&30.172&30.040&30.081&30.376\\
\hline
DnCNN&\textcolor{blue}{30.176}&33.059&30.871&\textcolor{blue}{29.405}&\textcolor{blue}{30.282}&\textcolor{red}{29.131}&29.432&32.438&29.996&30.211&\textcolor{blue}{30.096}&30.118&\textcolor{blue}{30.432}\\
\hline
FFDNet&30.062&\textcolor{red}{33.268}&30.786&29.331&30.142&29.052&29.431&\textcolor{red}{32.586}&29.978&\textcolor{blue}{30.226}&\textcolor{red}{30.101}&\textcolor{blue}{30.176}&30.428\\
\hline
Proposed&\textcolor{red}{30.233}&33.099&\textcolor{blue}{30.831}&\textcolor{red}{29.447}&\textcolor{red}{30.454}&\textcolor{blue}{29.111}&\textcolor{red}{29.462}&\textcolor{blue}{32.478}&30.063&\textcolor{red}{30.232}&30.091&\textcolor{red}{30.122}&\textcolor{red}{30.471}\\
\hline
Noise Level&\multicolumn{13}{|c|}{$\sigma = 50$}\\
\hline
BM3D&26.131&29.693&26.683&25.044&25.818&25.102&25.898&29.051&\textcolor{blue}{27.225}&26.782&26.808&26.463&26.722\\
\hline
WNNM&26.447&\textcolor{blue}{30.326}&26.941&25.442&26.312&25.413&26.132&29.244&\textcolor{red}{27.782}&26.968&26.938&26.635&27.025\\
\hline
TNRD&26.582&29.554&27.235&25.301&26.447&25.499&26.185&28.978&25.729&26.891&26.993&26.522&26.812\\
\hline
IRCNN&26.878&29.955&27.334&25.567&26.611&\textcolor{blue}{25.887}&\textcolor{blue}{26.521}&29.399&26.235&{27.173}&27.166&26.875&27.136\\
\hline
DnCNN&\textcolor{blue}{27.031}&30.002&27.321&25.701&26.781&25.865&26.483&29.385&26.217&27.201&\textcolor{blue}{27.242}&26.901&27.176\\
\hline
FFDNet&27.028&\textcolor{red}{30.432}&\textcolor{red}{27.428}&\textcolor{blue}{25.769}&\textcolor{blue}{26.882}&\textcolor{red}{25.901}&\textcolor{red}{26.576}&\textcolor{red}{29.677}&26.476&\textcolor{red}{27.318}&\textcolor{red}{27.296}&\textcolor{red}{27.068}&\textcolor{red}{27.321}\\
\hline
Proposed&\textcolor{red}{27.307}&30.177&\textcolor{blue}{27.415}&\textcolor{red}{25.781}&\textcolor{red}{26.931}&25.878&26.483&\textcolor{blue}{29.484}&26.417&\textcolor{blue}{27.287}&27.233&\textcolor{blue}{26.996}&\textcolor{blue}{27.282}\\
\hline
\end{tabular}
}
\label{table2}
\end{table*}

\section{Conclusion}
In this paper, we have designed an effective CNN denoisers for image denoising. Specifically, with the aid of skipped connections, we can easily train a deep and complex convolutional network. A lot of deep learning skills are integrated to speed up the training process and boost the denoising performance. The model-based and prior-based approaches are the popular way to tackle with denoising problem. Followed by the instruction of prior-based model, we show the possibility of increasing the features by using multiscale module and residual HDC module. Extensive experimental results have demonstrated that the proposed method can not only produce favorable image denoising performance quantitatively and qualitatively, but also have a promising run time by GPU. There are still some work for further study. First, it would be a promising direction to train a lightweight denoiser for practical applications. Second, extending the proposed method to other image restoration problems, such as image deblurring. Third, it would be interesting to investigate how to denoise the non-Gaussian noisy according to some properties of gaussian denoising models.

\begin{figure*}[t]
\centering
\includegraphics[width=17cm]{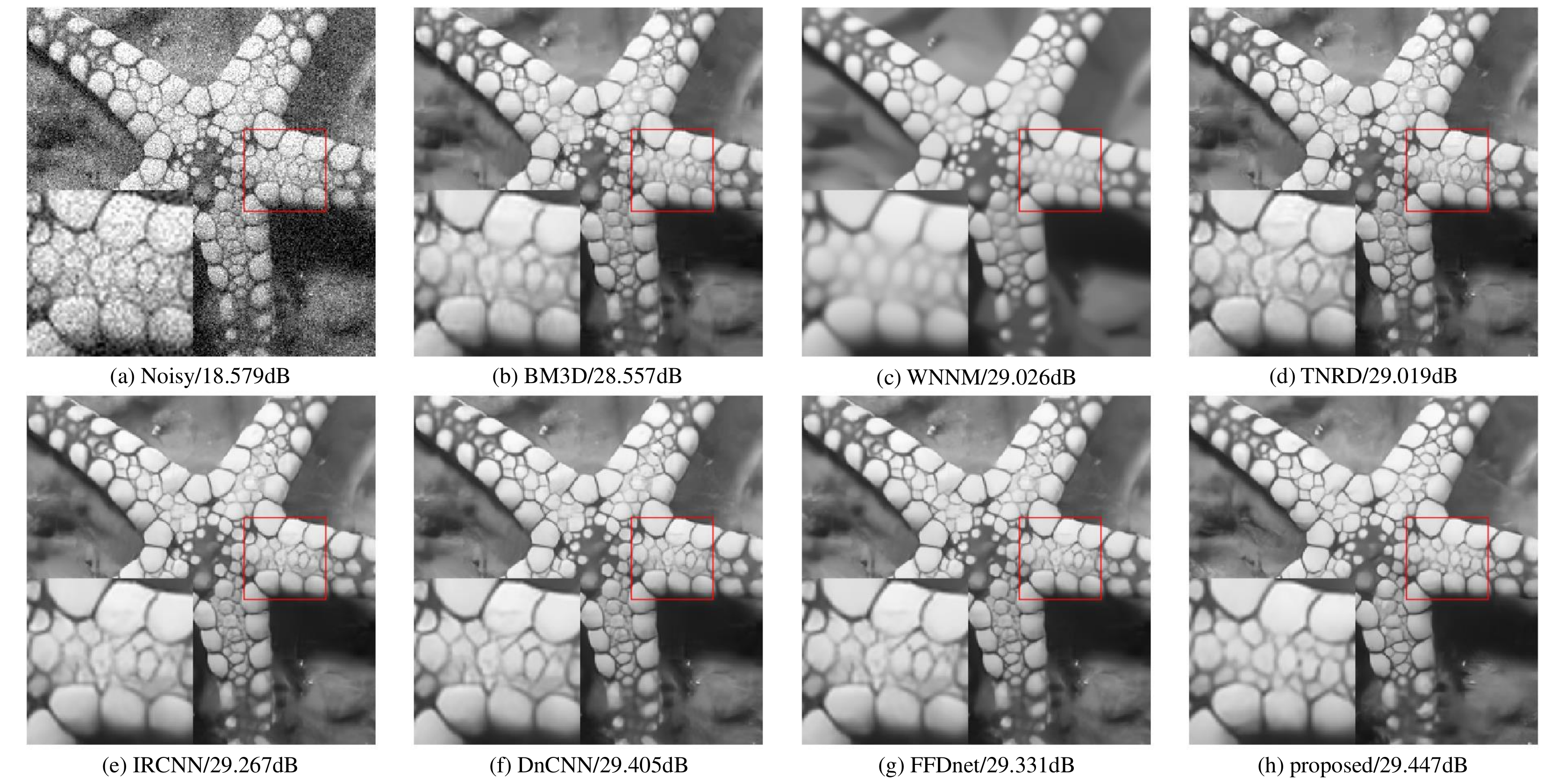}
\caption{Denoising results of image "Starfish" from Set12 with noise level 25}
\label{fig4}
\end{figure*}

\begin{figure*}[t]
\centering
\includegraphics[width=17cm]{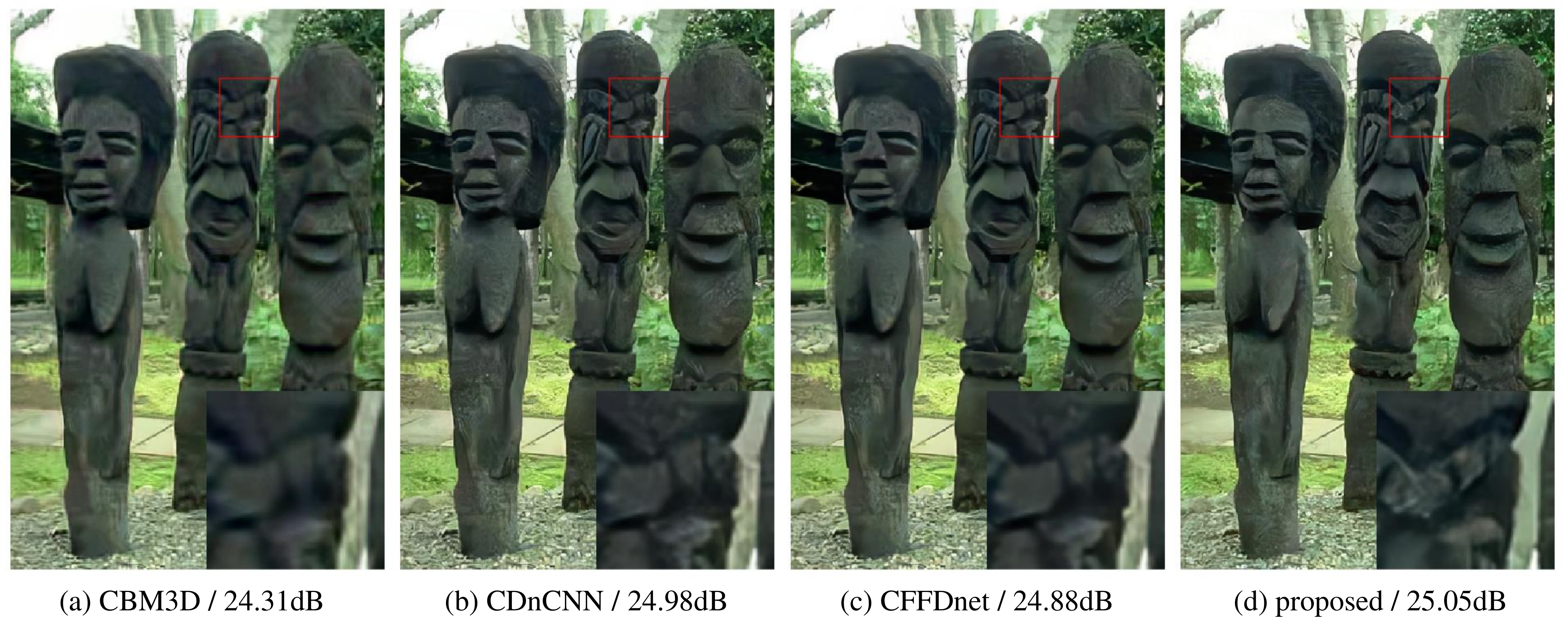}
\caption{Color image denoising results of one image from the CBSD68 dataset with noise level 50}
\label{fig6}
\end{figure*}




%




\end{document}